%% file: CatsvRNNsNeSyIfCoLog.tex
\documentclass[breaklinks]{myifcolog}
\usepackage{latexsym}

\usepackage{url}

\title{Compositionality for Recursive Neural Networks}
\titlerunning{Compositionality for TreeRNNs}

\addauthor[m.a.f.lewis@uva.nl]{Martha Lewis\thanks{Funded by NWO Veni grant `Metaphorical Meanings for Artificial Agents'}}
  {ILLC, University of Amsterdam}
 \authorrunning{Lewis}
 
 \titlethanks{Thanks to the organisers and participants of the NeSy2018 conference, and to the anoymous reviewers, for helpful comments and discussion.}

\usepackage{amsmath,amsthm,amssymb,stmaryrd}
\usepackage{xspace}
\usepackage[center]{qtree}
\usepackage{natbib}
\usepackage{setspace}
\usepackage{tikz, tikz-cd}
\usetikzlibrary{arrows,decorations.pathmorphing,backgrounds,positioning,fit}

\tikzset{every path/.style={thick}}
\tikzset{entry/.style={circle, thick, minimum size = 3mm, draw=blue!60, fill=blue!10}}
\tikzset{outer/.style={rectangle, thick, minimum height = 5mm, minimum width = 18mm, rounded corners = 7pt, draw=blue!80, fill = blue!20}}
\tikzset{
	pics/vector/.style={
		code={
			\draw (0,0) node [outer] {};
			\foreach \x in {-0.6, -0.2, 0.2, 0.6}
				\draw (\x, 0) node [entry] {};
		}
	}
}

\usepackage{hyperref}

\input{tikzpreamble}

\newcommand{\lang}[1]{\ensuremath{\textit{#1}}}
\newcommand{\langvec}[1]{\ensuremath{\overrightarrow{\lang{#1}}}}

\newcommand{\FVect}{\mathbf{FVect}}

\newcommand{\functor}[1]{\mathcal{#1}}

\newcommand{\vect}[1]{\overrightarrow{#1}}

\newcommand{\define}[1]{{\bf #1}}

\newcommand{\colvec}[2]{\begin{bmatrix} {#1} \\ {#2} \end{bmatrix}}

\theoremstyle{definition}\newtheorem{example}{Example}

\begin{document}
\maketitle
\begin{abstract} Modelling compositionality has been a longstanding area of research in the field of vector space semantics. The categorical approach to compositionality maps grammar onto vector spaces in a principled way, but comes under fire for requiring the formation of very high-dimensional matrices and tensors, and therefore being computationally infeasible. In this paper I show how a linear simplification of recursive neural tensor network models can be mapped directly onto the categorical approach, giving a way of computing the required matrices and tensors. This mapping suggests a number of lines of research for both categorical compositional vector space models of meaning and for recursive neural network models of compositionality.
\end{abstract}

\section{Introduction}
Vector space semantics represents the meanings of words as vectors, learnt from text corpora. In order to compute the meanings of multi-word phrases and sentences, the principle of compositionality is invoked. This is that for a sentence $s = w_1 w_2 ... w_n$ there should be a function $f_s$ that when applied to representations of the words $w_i$, will return a representation of the sentence $s$:
\[
s = f_s(w_1, w_2, ... w_n)
\]

One way to model meanings in a vector space is to use co-occurrence statistics \citep{Bullinaria2007}. The meaning of a word is identified with the frequency with which it appears near other words. A drawback of this approach is that antonyms appear in similar contexts and, hence are indistinguishable. Another related difficulty is that vector spaces are notoriously bad for representing basic propositional logic.  Nonetheless, the vector space model is highly successful in NLP. To model how words compose, a number of proposals have been made. These range from the simpler additive or multiplicative models given in \citet{Mitchell2010} to full-blown tensor contraction models \citep{CoeckeSadrzadehClark2010, maillard2014}. In between is the Practical Lexical Function model of \citet{paperno2014} which uses matrices to form function words such as adjectives and verbs.

The categorical compositional distributional model of \citet{CoeckeSadrzadehClark2010} implements compositionality by mapping each grammatical type to a corresponding vector space. Grammatical reductions between types are modelled as linear maps between these vector spaces. 
Well-typed sentences reduce to vectors in the sentence space $S$. Vectors for nouns are learnt using cooccurrence statistics in corpora. Adjectives and verbs can be learnt using multilinear regression \citep{BaroniZamparelli2010,GrefenstetteDinuZhangSadrzadehBaroni2013}, via a form of extensional composition \citep{GrefenstetteSadrzadeh2011}, or by using techniques that reduce the size of the vector space \citep{KartsaklisSadrzadehPulman2012}.

Another way of building word meanings is via neural embeddings \citep{mikolov2013}. This strategy trains a network to predict nearby words by maximizing the probability of observing words in the neighbourhood of another. This is similar to the distributional idea, but rather than counting words, they are predicted. The prediction can happen in two directions: either a word is predicted from its context, called the continuous bag-of-words model, or the context is predicted from the word, called the skip-gram model. This method can then be extended to give a notion of compositionality. Recursive neural networks as used in \citet{SocherPerelyginWuChuangManningNgPotts2013} and \citet{BowmanPotts2015} use a `compositionality function' that computes the combination of two word vectors. This pairwise combination is applied recursively in a way that follows the parse structure of the phrase. The compositionality function has the structure of a feedforward neural network layer, possibly with additions such as a tensor layer. The parameters for the compositionality function and for the vectors themselves are trained using backpropagation.

The categorical approach maps nicely to formal semantics approaches. The role of verbs and adjectives as functions from the noun space to other spaces is clearly delineated. Words such as relative pronouns, whose meanings are not well modelled by distributional approaches, can be given a purely mathematical semantics. However, the representations of functional words soon become extremely large, so that learning, storing, and computing with these representations becomes infeasible. Another difficulty with this framework is that word types are fixed, so that there is no easy way to move between, say, noun meanings and verb meanings.

Neural network approaches in general do not have an explicit connection with formal semantics. In the case of recursive neural networks there is some connection, since the structure of the network respects the parse structure, but there is limited consideration of different grammatical types and how these might be used. Different grammatical types are all represented within the same vector space. Words that arguably have more of an `information routing' function (such as pronouns, coordinators and so on) are also represented as vectors. However, these approaches are extremely successful. The word representations and the compositionality functions are more tractable than those of the full-blown tensor approach, and it is easy to consider a word vector as representing a number of different grammatical types - the same vector can be used to represent the noun `bank' in `financial bank' and the verb `bank' in `bank winnings'.

This paper shows how to understand a simplification of recursive neural networks within the categorical framework, namely, when the compositionality function is linear. Understanding recursive neural networks within this framework opens the door for us to use methods from formal semantics together with the neural network approach. I give an example of how we can express relative pronouns (words such as `who') and reflexive pronouns (`himself') within the framework. This mapping also benefits the categorical approach. The high-order tensors needed for the categorical approach can be dispensed with, and word types can be made more fluid. 

In the following, I firstly (section \ref{sec:disco}) give a description of categorical compositional vector space semantics. I go on to describe recursive neural networks and recursive neural tensor networks (section \ref{sec:NN}). In section \ref{sec:mapping} I show how linear recursive (tensor) networks can be given exactly the same structure as the categorical compositional model. Sections \ref{sec:benefits} and \ref{sec:conc} outline the benefits of this analysis for each approach, and discuss how we can take the analysis further. In particular the possibility of reintroducing the non-linearity in recursive neural networks is considered.

\section{Categorical Compositional Vector Semantics}
\label{sec:disco}
In this section I describe elements of the category-theoretic compositional approach to meaning, as given in \citet{CoeckeSadrzadehClark2010}, and show the general method by which the grammar category induces a notion of concept composition in the semantic category. An introduction to the kind of category theory used here is given in \citet{CoeckePaquette2011}.
The outline of the general programme is as follows \citep{bolt2017}:
\begin{enumerate}  
  \label{item:compstruct}
\item \begin{enumerate}
\item Choose a compositional structure, such as a categorial grammar.
\item Interpret this structure as a category, the \define{grammar category}.
\end{enumerate}
\item \begin{enumerate}
  \label{item:meaningspace}
\item Choose or craft appropriate meaning or concept spaces, such as vector spaces.
  \label{item:meaningcategory}
\item Organize these spaces into a category, the \define{semantics category}, with the same abstract structure as the grammar category.
\end{enumerate}
  \label{item:interpret}
\item Interpret the compositional structure of the grammar category in the semantics category via a functor preserving the type reduction
  structure.
  \label{item:reduction}
\item This functor maps type reductions in the grammar category onto algorithms for composing meanings in the semantics category. 
\end{enumerate}

This paper describes one instantiation of this approach, using pregroup grammar and the category $\FVect$ of vector spaces and linear maps. This paper will use pregroup grammar, but it is also possible to use other approaches such as other categorial grammars, described in \citet{Coecke2013}.

\subsection{Pregroup grammar}
The description of pregroup grammars given follows that of \citet{PrellerSadrzadeh2011}. Whilst the details are slightly technical, the form of the grammar is very intuitive. Essentially we require a category that has types for nouns and for sentences, together with adjoint types, which are similar to inverses, a method for concatenating them, and morphisms that correspond to type reductions. The structure we desire for this category is termed \emph{compact closed}. Details are given in \citet{CoeckeSadrzadehClark2010} and  \citet{PrellerSadrzadeh2011}.

The category $G$ used for grammar is roughly as follows. The grammar is built over a set of types. We consider the set containing just $n$ for noun and $s$ for sentence. Each type has two adjoints $x^r$ and $x^l$. Complex types can be built up by concatenation of types, for example $x\cdot y^l\cdot z^r$, and we often leave out the dot so $xy = x\cdot y$. There is also a unit type such that $x1 = 1x = x$. Types and their adjoints interact via the following morphisms:
  \begin{align*}
    \epsilon^r_x:x\cdot x^r \rightarrow 1, &\qquad \epsilon^l_x:x^l\cdot x\rightarrow 1\\
    \eta^r_x:1 \rightarrow x^r\cdot x, &\qquad
    \eta^l_x:1\rightarrow x\cdot x^l
  \end{align*}
The morphisms $\epsilon^r_x$ and $\epsilon^l_x$ can be thought of as \emph{type reduction} and the morphisms $\eta^r_x$ and $\eta^l_x$ can be thought of as \emph{type introduction}. A string of grammatical types $t_1, ... t_n$ is then said to be grammatical if it reduces, via the morphisms above, to the sentence type $s$.

\begin{example}
\label{ex:pregred}
Consider the sentence \textit{`dragons breathe fire'}. The nouns \textit{dragons} and \textit{fire} are of type $n$, and the verb \textit{breathe} is given the type $n^r s n^l$. \textit{`dragons breathe fire'} therefore has type~$n(n^rsn^l)n$. Then we have the following type reductions:
\begin{align*}
  (\epsilon^r_n \cdot id_s \cdot \epsilon ^l_n)( n(n^rsn^l)n) &= (\epsilon^r_n \cdot id_s \cdot \epsilon ^l_n)((nn^r)s(n^ln))\\
  & \rightarrow (id_s \cdot \epsilon ^l_n) s(n^ln) \rightarrow s 
\end{align*}
The above reduction can be given a neat graphical interpretation as follows:
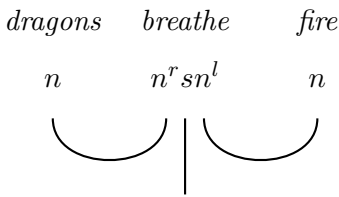
\begin{center}
\input{chickens.tikz}  
\end{center}
This diagrammatic calculus is fully explained in \citet{CoeckeSadrzadehClark2010}, amongst others. Essentially we can think of the u-shaped `cups' as type reductions, and calculations can be made by manipulating the diagrams as if they lie on a flat plane, maintaining numbers of inputs and outputs.
\end{example}

\subsection{Mapping to vector spaces}
We use the category $\FVect$ of finite dimensional vector spaces and linear maps, which is also compact closed.  We describe a functor $\functor{F}:G \rightarrow \FVect$ that maps the noun type $n$ to a vector space $N$, the sentence type $s$ to $S$, the unit $1$ to $ \mathbb{R}$, concatenation maps to $\otimes$, i.e., the tensor product of vector spaces, adjoints are lost, 
$\epsilon^r_p$ and  $\epsilon^l_p$ map to tensor contraction, and 
$\eta^r_p$ and $\eta^l_p$ map to identity maps.
\begin{example}
Consider again the sentence \textit{`dragons breathe fire'}. The nouns \textit{dragons} and \textit{fire} have type $n$ and so are represented in some vector space $N$ of nouns. The transitive verb \textit{breathe} has type $n^r s n^l$ and, hence, is represented by a vector in the vector space $N\otimes S \otimes N$ where $S$ is a vector space modelling sentence meaning. The meaning of \textit{`dragons breathe fire'} is the outcome of applying the type reduction morphisms given in 
\begin{equation}\label{eq:pgex2}
\epsilon_N \otimes 1_S \otimes \epsilon_N : N\otimes ( N\otimes S\otimes N ) \otimes N \rightarrow S
\end{equation}
i.e. sequences of tensor contractions, to the product
\begin{equation}\label{eq:pgex1}
\vect{\mathit{dragons}}\otimes\vect{\mathit{breathe}}\otimes\vect{\mathit{fire}}
\end{equation}
\end{example}
This nicely illustrates the general method. The meaning category supplies vectors for \textit{dragons}, \textit{breathe}, and \textit{fire}. The grammar category then tells us how to stitch these together. The essence of the method should be thought of as the diagram
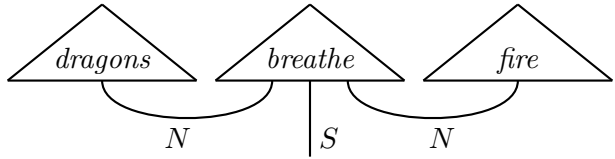
\begin{center}
\input{chickens2.tikz}
\end{center}
where we think of the words as meaning vectors (\ref{eq:pgex1}) and the wires as the map (\ref{eq:pgex2}). Again, the `cups' can be thought of as type reductions. Linear-algebraically, the map (\ref{eq:pgex2}) and the diagram above are equivalent to the following. Suppose we have a set of basis vectors $\{\vect{e}_i\}_i$. Define 
\[\vect{\lang{dragons}}  = \sum_i d_i \vect{e}_i, \qquad
\vect{\lang{breathe}} = \sum_{ijk} b_{ijk} \vect{e}_i \otimes\vect{e}_j \otimes\vect{e}_k, \qquad
\vect{\lang{fire}}  = \sum_i f_i \vect{e}_i\]
Then 
\begin{align*}
&\vect{\lang{dragons breathe fire}} = (\epsilon_N \otimes 1_S \otimes \epsilon_N) \vect{\mathit{dragons}}\otimes\vect{\mathit{breathe}}\otimes\vect{\mathit{fire}}\\
&\qquad= (\epsilon_N \otimes 1_S \otimes \epsilon_N)\left(\sum_i d_i \vect{e}_i \otimes
\sum_{jkl} b_{jkl} \vect{e}_j \otimes\vect{e}_k \otimes\vect{e}_l \otimes \sum_m f_m \vect{e}_m\right)\\
&\qquad= (1_S \otimes \epsilon_N)\left(\sum_{ijkl} d_i b_{jkl} \vect{e}_j \otimes\vect{e}_k \otimes \sum_m f_m \vect{e}_m\right) = \sum_{ijk} d_i b_{ijk} f_k\vect{e}_j
\end{align*}
where this last expression is a single vector in the sentence space.
\section{Neural Network Models}
\label{sec:NN}
Neural networks are used both as a way of building meaning vectors and as a way of modelling compositionality in meaning spaces. \citet{mikolov2013} describes a pair of methods that build vectors by using context windows, and making predictions about the likely content of either the context window or the word itself. Phrases and sentences are represented in the same space as the words. To compute vectors for multi-word sentences and phrases, \citet{SocherPerelyginWuChuangManningNgPotts2013} use tree-structured recursive neural networks. The phrases and sentences output by the network can then be used for various tasks, notably sentiment analysis. The sections below summarise recursive neural networks and recursive tensor neural networks. In the following sections we assume that words are represented as vectors in $\mathbb{R}^n$.

\subsection{Recursive neural networks}
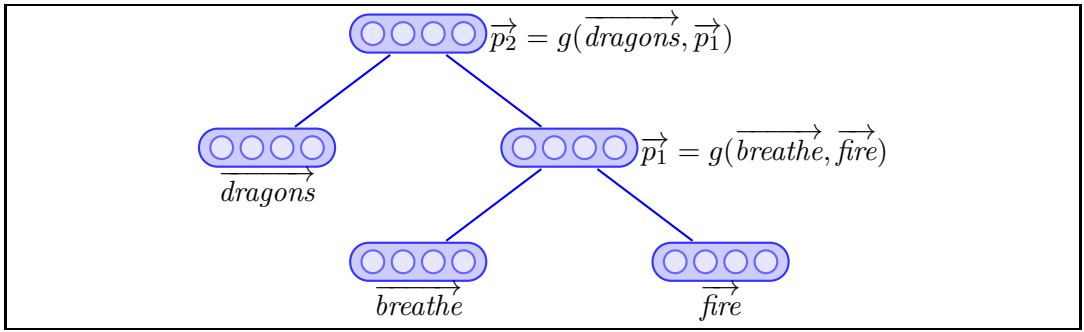
\begin{figure}
\centering
\begin{tikzpicture}
[edge from parent, sibling distance=40mm, level distance=15mm,
   edge from parent/.style={blue,thick,draw}, every node/.style={inner sep = 0pt}]
\node[label=right:{$\vect{p_2} = g(\langvec{dragons}, \vect{p_1})$}] {\tikz{\pic{vector}}}
child {node[label=below:{\langvec{dragons}}]{\tikz{\pic{vector}}}}
child {node[label=right:{$\vect{p_1} = g(\langvec{breathe}, \langvec{fire})$}]{\tikz{\pic{vector}}}
child {node[label=below:{\langvec{breathe}}]{\tikz{\pic{vector}}}}
child {node[label=below:{\langvec{fire}}]{\tikz{\pic{vector}}}}
};
\end{tikzpicture}
\caption{Schematic of an TreeRNN. Word vectors and/or parent vectors $p_i$ are combined using the compositionality function $g$ according to the parse tree. The vector $\vec{p_1}$ corresponds to the verb phrase \lang{breathe fire} and the vector $\vec{p_2}$ corresponds to the whole sentence \lang{dragons breathe fire}.}
\label{fig:RNNtree}
\end{figure}

Recursive neural networks (TreeRNNs) have a tree-like structure. When applied to sentences, the tree represents the syntactic structure of the sentence.
A schematic of a recursive neural network is given in Figure \ref{fig:RNNtree}. The words of a sentence are represented as vectors. Words can be combined via the \emph{compositionality function} $g$ to form a parent vector. In the networks we discuss here, the parent vectors are of the same dimensionality as the input vectors, meaning that the compositionality function can be applied recursively according to the parse tree. The compositionality function and the input vectors themselves are learnt by error backpropagation.

The compositionality function for a TreeRNN is usually of the form
\[
g_{\textit{TreeRNN}}:\mathbb{R}^n \times \mathbb{R}^n \rightarrow \mathbb{R}^n :: (\vect{v_1}, \vect{v_2}) \mapsto f_1\left(M\cdot\colvec{\vect{v_1}}{\vect{v_2}}\right)
\]

where $\vect{v_i} \in \mathbb{R}^n$, $\colvec{-}{-}$ is vertical concatenation of column vectors, $M \in \mathbb{R}^n \otimes \mathbb{R}^{2n}$, and $(-\cdot-)$ is tensor contraction. $f_1$ is a squashing function that is applied pointwise to its vector argument, for example $f = \tanh$. The parent vector that forms the root of the tree is the representation of the whole sentence. Parent vectors within the tree represent subphrases of the sentence. The matrix $M$ and the input vectors are learnt during training.

\subsection{Recursive neural tensor networks}
Recursive neural tensor networks (TreeRNTNs) are similar to TreeRNNs but differ in the compositionality function $g$. The function $g$ is as follows:
\[
g_{\textit{TreeRNTN}}:\mathbb{R}^n \times \mathbb{R}^n \rightarrow \mathbb{R}^n :: (\vect{v_1}, \vect{v_2}) \mapsto g_{TreeRNN}(\vect{v_1}, \vect{v_2}) +  f_2\left(\vect{v_1}^\top\cdot T \cdot\vect{v_2}\right)
\]

where $\vect{v_i}$ and $(-\cdot-)$ are as before, $T \in \mathbb{R}^n\otimes \mathbb{R}^n \otimes \mathbb{R}^n$ and $f_2$ is a squashing function. Again, the input vectors, matrix $M$ and tensor $T$ are learnt during training.

\section{Mapping between categorical and TreeRNN compositionality}
\label{sec:mapping}
It is now possible to model a simplifed version of TreeRNNs within the categorical vector space semantics of \citet{CoeckeSadrzadehClark2010}. I show show how a linearized version can be modelled within $\FVect$ using pregroup grammar as the grammar category.

With a (drastic) simplification of the compositionality function $g_{\textit{TreeRNTN}}$ there is an immediate correspondence between the TreeRNTN model and a simplified version of the categorical model. We drop both the non-linearity and the matrix part of the function $g$, giving:

\[
g_{\textit{Lin}}:\mathbb{R}^n \times \mathbb{R}^n \rightarrow \mathbb{R}^n :: (\vect{v_1}, \vect{v_2})  \mapsto \left(\vect{v_1}^\top\cdot T \cdot\vect{v_2}\right)
\]


Now the tensor $T$ is just a multilinear map, i.e., morphism in $\FVect$, and we can therefore describe a direct translation between linear TreeRNTNs and categorical compositional vector space semantics with pregroups.

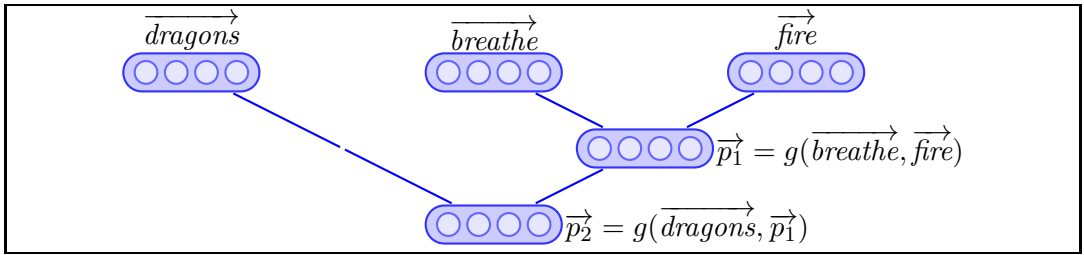
\begin{figure}[h!]
\centering
\begin{tikzpicture}
[grow'=up, sibling distance=40mm, level distance=10mm, edge from parent/.style={blue,thick,draw}, every node/.style={inner sep = 0pt}]
\node[label=right:{$\vect{p_2} = g(\langvec{dragons}, \vect{p_1})$}] {\tikz{\pic{vector}}}
child {node {}
child {node[label=above:{\langvec{dragons}}]{\tikz{\pic{vector}}}}
child {node {} edge from parent[color=white]}}
child {node[label=right:{$\vect{p_1} = g(\langvec{breathe}, \langvec{fire})$}]{\tikz{\pic{vector}}}
child {node[label=above:{\langvec{breathe}}]{\tikz{\pic{vector}}}}
child {node[label=above:{\langvec{fire}}]{\tikz{\pic{vector}}}}
};
\end{tikzpicture}
\caption{The TreeRNN schematic turned upside down and one edge lengthened}
\label{fig:RNNtreeflipped}
\end{figure}

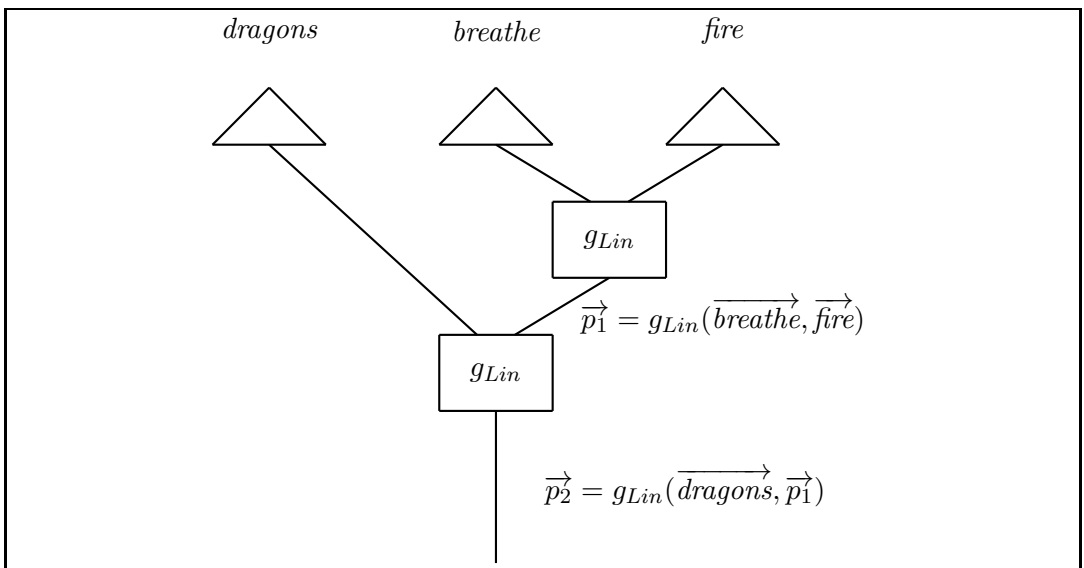
\begin{figure}[h!]
\centering
\input{RNN.tikz}
\caption{The schematic translated into the diagrammatic calculus. The compositionality function $g_{Lin}$ is just a tensor with no nonlinearity applied.}
\label{fig:RNNdia}
\end{figure}

\begin{figure}[h!]
\centering
\input{RNN2.tikz}
\caption{The diagram in figure \ref{fig:RNNdia} with wires bent. This is allowed since we are now working in the category $\FVect$.}
\label{fig:RNNbent}
\end{figure}
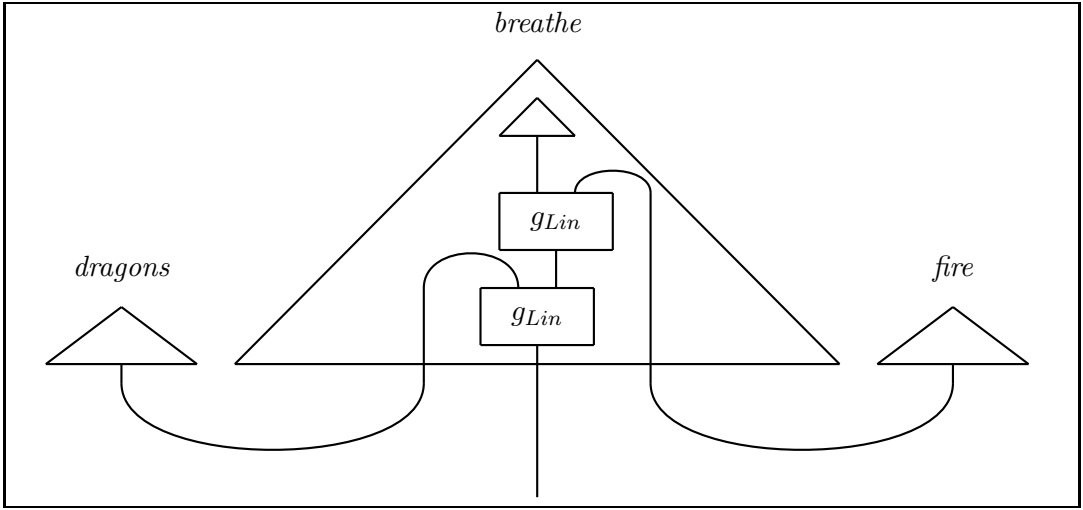

Recall that in the categorical model we had a nice diagrammatic calculus to represent the calculations we were making. We also had a schematic for the TreeRNNs. With the simplified compositionality function, we can translate that schematic into the diagrammatic calculus, shown in figures \ref{fig:RNNtreeflipped}, \ref{fig:RNNdia}, and \ref{fig:RNNbent}.

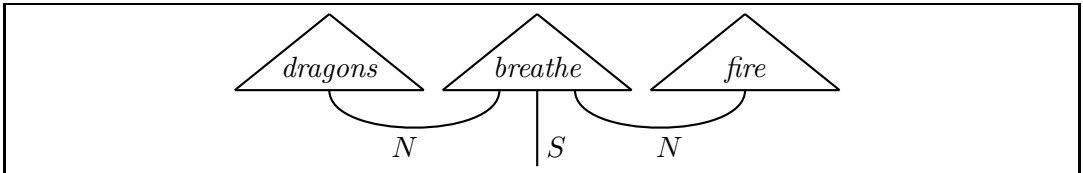
\begin{figure}[h!]
\centering
\input{chickens2.tikz}
\caption{We can therefore see the case in \ref{fig:RNNbent} as an instance of the categorical method, where the interior of the tensor is created using two instances of the compositionality function $g_{Lin}$}
\end{figure}

\begin{figure}[h!]
\centering
\input{RNNA.tikz}
\caption{Adjective formed from part of a TreeRNTN.}
\label{fig:RNNA}
\end{figure}
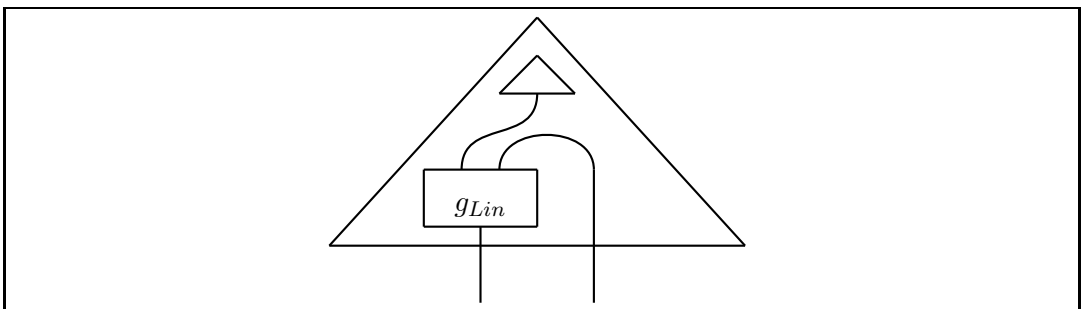

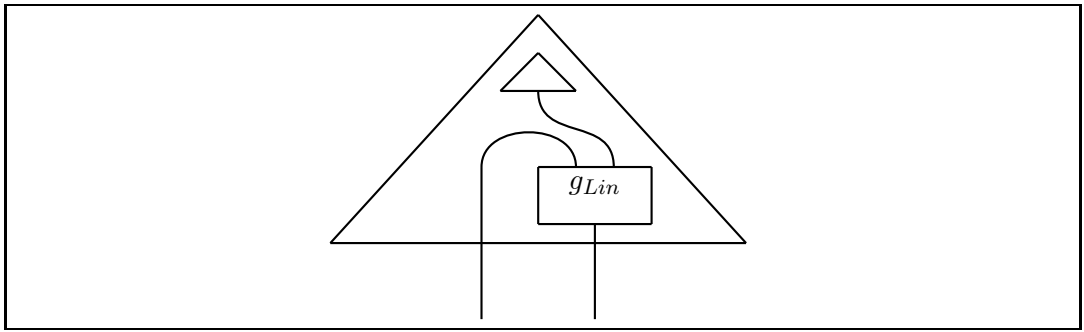
\begin{figure}[h!]
\centering
\input{RNNIV.tikz}
\caption{Intransitive verb formed from part of an TreeRNTN.}
\label{fig:RNNIV}
\end{figure}

These diagrams show how the interior of the verb has been has been analysed into two instances of the compositionality function wired together, with the verb vector $\langvec{breathe}$ as input. This means that rather than learn large numbers of parameters for each word in the lexicon, just one tensor comprising the compositionality function needs to be learnt, together with vectors in $N$ for each word. This mapping can be carried out for other parts of speech. The representations of adjectives and intransitive verbs are given in figures \ref{fig:RNNA} and \ref{fig:RNNIV}, each requiring just one instance of the compositionality function. In section \ref{sec:RNNbens}, we discuss how we can analyze other sorts of words such as relative pronouns and reflexive pronouns.

\section{Benefits}
\label{sec:benefits}
In outlining this comparison a number of benefits arise. This section outlines benefits for the categorical model and then for RNN models.

\subsection{Categorical models}
One of the main charges levelled at the categorical compositional distributional semantics mode is that the dimensionality of the tensors required is too high, and that training is too expensive. The correspondence I have outlined here gives an approach where the number of high-dimensional tensors to train is limited. 

In the simplest case, one linear compositionality function could be learned, together with vectors for each word. The learning algorithm for this approach will be similar to strategies used for training recursive neural networks. The networks will therefore be as easy, or easier, to train than the TreeRNNs used by \citet{SocherPerelyginWuChuangManningNgPotts2013} and \citet{BowmanPotts2015}. However, since the compositionality functions to train are now linear, the results obtained are unlikely to be as good as those obtained using full TreeRNNs. One strategy to alleviate this is as follows. Different compositionality functions could be used for different word types. So, for example, we would have functions $g_{\lang{adj}}$ for an adjective, $g_{\lang{iv}}$ and $g_{\lang{tv}}$ for a transitive verb. For example, the functions for the adjective is $g_{\lang{adj}}(\vect{v}_n) = \vect{v}_a^\top T_{\lang{adj}} \vect{v}_n$, and for an intransitive verb is $g_{\lang{iv}}(\vect{v}_n) = \vect{v}_n^\top T_{\lang{iv}} \vect{v}_i$, 
where $\vect{v}_a$ is the vector of the adjective, $\vect{v}_i$ is the vector of the intransitive verb, and $\vect{v}_n$ is the vector of the noun. Using this strategy, the noun space and the sentence space can now be separated so that sentences no longer have to inhabit the same space as nouns.

A further benefit for the categorical model is that this approach alleviates the brittleness of the representations learnt. Rather than learning individual tensors for each functional word, we are simply learning a small number of compositionality functions. This means that we can switch between the noun `bank' and the verb `bank' simply by plugging the word vector $\lang{bank}$ into the relevant function.

Furthermore, since this approach is a simplification of the model of \citet{CoeckeSadrzadehClark2010}, extensions of that model can also be applied.  In particular, information-routing words like relative pronouns can  be modelled using the approaches outlined in \citet{FrobMeanI}. This is discussed further in the next section.

\subsection{TreeRNN models}
\label{sec:RNNbens}
Although TreeRNNs have fewer parameters and more flexibility than the categorical vector space models, the compositional mechanism they use is `one size fits all'. The TreeRNN approach as elaborated so far does not distinguish between content words such as `dog',`brown', and information routing words such as pronouns and logical words.
The approach outlined here makes an explicit connection between formal semantics approaches in the form of pregroup grammars on the one hand, and neural network approaches for composition on the other. This means that we can use strategies from formal semantics to represent the meaning of information routing words. The benefit of doing so is two-fold. Firstly, it may improve training time, since the compositionality function will not have to encompass this aspect of composition. Secondly, by separating out some of the compositional mechanism, we make the behaviour of the network more transparent.  The roles of certain words will be modelled as functions that do not need to be learnt. I give below two examples: relative pronouns as analysed in \citet{FrobMeanI} and reflexive pronouns.

\subsubsection{Relative pronouns}
\citet{FrobMeanI} analyze relative pronouns by using the Frobenius algebra structure available on finite-dimensional vector spaces. Full details of how Frobenius algebras are defined and used are given in those papers, but briefly, we can consider these to introduce copying, merging, and deleting mechanisms into the semantics.  

In $\FVect$, any vector space $V$ with a fixed basis $\{\vect{e_i}\}_i$ has a Frobenius algebra over it, explicitly given by:
\begin{align}
&\Delta:: \vect{e_i} \mapsto \vect{e_i} \otimes \vect{e_i}  &\iota :: \vect{e_i} \mapsto 1 \\
&\mu:: \vect{e_i} \otimes \vect{e_i} \mapsto \vect{e_i}  &\zeta :: 1 \mapsto \vect{e_i}
\end{align}
Linear-algebraically, the $\Delta$ morphism takes a vector and embeds it into the diagonal of a matrix. The $\mu$ morphism takes a matrix $z \in W \otimes W$ and returns a vector consisting only of the diagonal elements of $z$. If the matrix $z$ is the tensor product of two vectors $z = 
\vect{v} \otimes \vect{w}$, then $\mu(v \otimes w) = v \odot w$ where $(-\odot -)$ corresponds to pointwise multiplication. These operations extend to higher-order tensors.

In pregroup grammar, the word `who' is given the type $n^r n s^l n$. Rather than learn parameters for an order 4 tensor, \citet{FrobMeanI} show how it can be given a purely mathematical meaning. This is shown diagrammatically below:
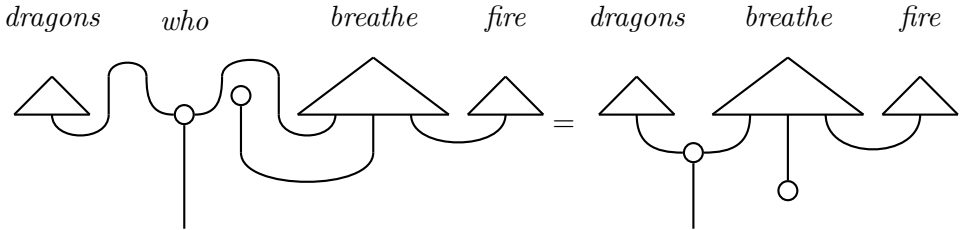
\[
\begin{gathered}\input{relpron.tikz}\end{gathered} = \begin{gathered}\input{relpronyanked.tikz}\end{gathered} 
\]
The word `who' is equivalent to discarding the sentence part of the verb and pointwise multiplying the vectors for \lang{dragons} and \lang{breathe fire}.

\subsubsection{Reflexive Pronouns}
Reflexive pronouns are words such as `himself'. These words also have an information routing role. In a sentence like \lang{John loves himself}, we want the content of \lang{John} to be copied out and routed to the object of the verb. The pregroup type of the pronoun `himself' can be given as $n s^r n^{rr}n^r s$. We can give the reflexive pronoun a purely mathematical semantics as follows:

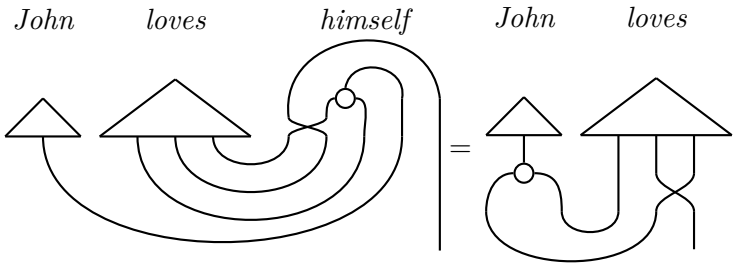
\[
\begin{gathered}\input{reflexivepron.tikz}\end{gathered} = \begin{gathered}\input{reflexivepronyanked.tikz}\end{gathered} 
\]

The reflexive pronoun takes in the noun, copies it, and plugs it into both the subject and the object of the verb, and returns the resulting sentence.

This treatment of reflexive and relative pronouns is part of a larger programme, relating vector space models of meaning and formal semantics. The idea is that some words can be thought of as `information routing' - they move information around a sentence, and at least part, if not all, of their meaning should be purely mathematical. In contrast, information-carrying words like nouns and adjectives, have meaning determined by co-occurrence, rather than by a mathematical function. In the TreeRNN approach, this distinction is not made, meaning that the compositionality function learnt must take into account both statistical and information-routing kinds of meaning. The proposal here is that information-routing words can be understood as part of the structure of the tree, rather than as vectors.

\section{Conclusions and Further Work}
\label{sec:conc}
The aim of this paper is to set out a mapping between the categorical compositional vector space semantics of \citet{CoeckeSadrzadehClark2010} and the recursive neural network (TreeRNN) models of \citet{SocherPerelyginWuChuangManningNgPotts2013} and \citet{BowmanPotts2015}. I have shown how a linear version of TreeRNNs can be modelled directly within the categorical model. This gives a strategy for simplifying the training for the categorical model, and also means that the categorical model is more flexible in its word representations. Since a linearized neural network is not going to be as successful as a standard network, I have also suggested learning individual networks for individual grammatical types, as a way of improving performance whilst still requiring many fewer parameters than the standard categorical model.
Modelling TreeRNNs within the categorical framework means that we can use ideas from formal semantics to simplify networks. I have shown how relative pronouns and reflexive pronouns can be analysed as having a purely mathematical semantics. This means that the networks learnt do not need to take this sort of compositionality into account. Furthermore, using the purely mathematical semantics when available means that the networks are more transparent. With analysis of these words, the compositionality function learnt can specialise to contentful words, rather than information routing words.

\subsection{Further work}
\label{sub:nonlin}
Section \ref{sec:mapping} showed how we can express a linear version of TreeRNTNs within the categorical compositional vector space model. However, using only linear transformations limits what these networks can do. Ongoing work is to examine how non-linearity can be reintroduced, by changing the categorical framework in which we work. The most promising avenue seems to be to change to monoidal biclosed categories and Lambek categorial grammar.

There are a number of other avenues for further work to be considered. On the implementation side: 
\begin{itemize}
\item The performance of linear TreeRNNs can be tested against the usual categorical apporaches to learning words.
\item The performance of linear TreeRNNs with specialised word-type networks can be tested against standard TreeRNNs.
\item The performance of TreeRNNs with formally analyzed information-routing words can be tested.
\item The effects of switching between word types can be investigated.
\end{itemize}
On the theoretical side:
\begin{itemize}
\item The analysis of reflexive pronouns can be extended to other pronouns and anaphora.
\item Investigating meanings of logical words and quantifiers.
\item Extending the analysis to other types of recurrent neural network such as long short-term memory networks or gated recurrent units.
\end{itemize}

\bibliographystyle{plainnat}
\bibliography{cs.bib}

\end{document}

%% file: tikzpreamble.tex
\usepackage{tikz}
\usepackage{pgfplots}
\usetikzlibrary{trees}
\usetikzlibrary{topaths}
\usetikzlibrary{decorations.pathmorphing}
\usetikzlibrary{decorations.markings}
\usetikzlibrary{matrix,backgrounds,folding}
\usetikzlibrary{chains,scopes,positioning,fit}
\usetikzlibrary{arrows,shadows}
\usetikzlibrary{calc} 
\usetikzlibrary{chains}
\usetikzlibrary{shapes,shapes.geometric,shapes.misc}

\pgfdeclarelayer{edgelayer}
\pgfdeclarelayer{nodelayer}
\pgfsetlayers{background,edgelayer,nodelayer,main}

\tikzstyle{dot}=[circle,fill=black,draw=black]
\tikzstyle{none}=[inner sep=0pt]
\tikzstyle{every loop}=[]

\tikzstyle{(null)}=[]
\tikzstyle{plain}=[]

\tikzstyle{blank}=[inner sep=0pt]
\tikzstyle{box}=[rectangle, minimum size = 0.5cm,fill=white,draw=black]
\tikzstyle{small_node}=[inner sep=0pt, minimum size=0.2cm,circle,fill=white,draw=black]
\tikzstyle{square}=[thick, minimum size=0.3cm,rectangle,fill=white,draw=black]
\tikzstyle{small_circle}=[inner sep=0pt, minimum size=0.25cm,circle,fill=white,draw=black]
\tikzstyle{big_circle}=[inner sep=0pt, minimum size=2cm,circle,fill=none,draw=black]
\tikzstyle{big_rectangle}=[inner sep=0pt, thick, rounded corners, minimum height=2cm, minimum width=2.5cm,rectangle,fill=none,draw=black]
\tikzstyle{circle_node}=[inner sep=0pt, minimum size=0.5cm,circle,fill=white,draw=black]
\tikzstyle{medium_circle}=[inner sep=0pt, minimum size=1.5cm,circle,fill=none,draw=black]
\tikzstyle{vbig_rectangle}=[inner sep=0pt, thick, rounded corners, minimum height=5.5cm, minimum width=5.5cm,rectangle,fill=none,draw=black]

\tikzstyle{downtri}=[regular polygon,regular polygon sides=3,shape border rotate=180,fill=white,draw=black]
\tikzstyle{uptri}=[regular polygon,regular polygon sides=3,shape border rotate=0,fill=white,draw=black]

\tikzstyle{morph}=[->,draw=black,line width=0.600]
\tikzstyle{arrow}=[-,draw=black,postaction={decorate},decoration={markings,mark=at position .5 with {\arrow{>}}},line width=2.000]
\tikzstyle{tick}=[-,draw=black,postaction={decorate},decoration={markings,mark=at position .5 with {\draw (0,-0.1) -- (0,0.1);}},line width=2.000]

%% file: chickens.tikz
\begin{tikzpicture}
	\begin{pgfonlayer}{nodelayer}
		\node [style=none] (0) at (0, -0) {};
		\node [style=none] (1) at (-0.25, -0) {};
		\node [style=none] (2) at (0.25, -0) {};
		\node [style=none] (3) at (1.75, -0) {};
		\node [style=none] (4) at (-1.75, -0) {};
		\node [style=none, anchor=mid] (5) at (-1.75, 0.5) {$n$};
		\node [style=none, anchor=mid] (6) at (0, 0.5) {$n^r s n^l $};
		\node [style=none, anchor=mid] (7) at (1.75, 0.5) {$n$};
		\node [style=none] (8) at (0, -1) {};
		\node [style=none, anchor=mid] (9) at (-1.75, 1.25) {\lang{dragons}};
		\node [style=none, anchor=mid] (10) at (0, 1.25) {\lang{breathe}};
		\node [style=none, anchor=mid] (11) at (1.75, 1.25) {\lang{fire}};
	\end{pgfonlayer}
	\begin{pgfonlayer}{edgelayer}
		\draw (0.center) to (8.center);
		\draw [bend right=90, looseness=1.25] (4.center) to (1.center);
		\draw [bend right=90, looseness=1.25] (2.center) to (3.center);
	\end{pgfonlayer}
\end{tikzpicture}

%% file: chickens2.tikz
\begin{tikzpicture}[{every path/.style}=thick]
	\begin{pgfonlayer}{nodelayer}
		\node [style=none, anchor=mid] (0) at (-2.75, 1) {\lang{dragons}};
		\node [style=none, anchor=mid] (1) at (0, 1) {\lang{breathe}};
		\node [style=none, anchor=mid] (2) at (2.75, 1) {\lang{fire}};
		\node [style=none] (3) at (-2.75, 0.75) {};
		\node [style=none] (4) at (-0.5, 0.75) {};
		\node [style=none] (5) at (0, 0.75) {};
		\node [style=none] (6) at (0.5, 0.75) {};
		\node [style=none] (7) at (2.75, 0.75) {};
		\node [style=none] (8) at (0, -0.25) {};
		\node [style=none] (9) at (4, 0.75) {};
		\node [style=none] (10) at (1.5, 0.75) {};
		\node [style=none] (11) at (2.75, 1.75) {};
		\node [style=none] (12) at (-2.75, 1.75) {};
		\node [style=none] (13) at (-4, 0.75) {};
		\node [style=none] (14) at (-1.5, 0.75) {};
		\node [style=none] (15) at (0, 1.75) {};
		\node [style=none] (16) at (-1.25, 0.75) {};
		\node [style=none] (17) at (1.25, 0.75) {};
		\node [style=none] (18) at (-1.75, 0) {$N$};
		\node [style=none] (19) at (1.75, 0) {$N$};
		\node [style=none] (20) at (0.25, 0) {$S$};
	\end{pgfonlayer}
	\begin{pgfonlayer}{edgelayer}
		\draw [bend right=90, looseness=0.75] (3.center) to (4.center);
		\draw [bend right=90, looseness=0.75] (6.center) to (7.center);
		\draw (5.center) to (8.center);
		\draw [style=swap] (11.center) to (9.center);
		\draw [style=swap] (9.center) to (10.center);
		\draw [style=swap] (10.center) to (11.center);
		\draw [style=swap] (12.center) to (14.center);
		\draw [style=swap] (14.center) to (13.center);
		\draw [style=swap] (13.center) to (12.center);
		\draw [style=swap] (15.center) to (17.center);
		\draw [style=swap] (17.center) to (16.center);
		\draw [style=swap] (16.center) to (15.center);
	\end{pgfonlayer}
\end{tikzpicture}

%% file: RNN.tikz
\begin{tikzpicture}[{every path/.style}=thick]
	\begin{pgfonlayer}{nodelayer}
		\node [style=none] (0) at (-3, 2.75) {};
		\node [style=none] (1) at (-3.75, 2) {};
		\node [style=none] (2) at (-2.25, 2) {};
		\node [style=none] (3) at (0, 2.75) {};
		\node [style=none] (4) at (-0.75, 2) {};
		\node [style=none] (5) at (0.75, 2) {};
		\node [style=none] (6) at (3, 2.75) {};
		\node [style=none] (7) at (2.25, 2) {};
		\node [style=none] (8) at (3.75, 2) {};
		\node [style=none] (9) at (-3, 3.5) {\lang{dragons}};
		\node [style=none] (10) at (0, 3.5) {\lang{breathe}};
		\node [style=none] (11) at (3, 3.5) {\lang{fire}};
		\node [style=none] (12) at (0.75, 1.25) {};
		\node [style=none] (13) at (2.25, 1.25) {};
		\node [style=none] (14) at (0.75, 0.25) {};
		\node [style=none] (15) at (2.25, 0.25) {};
		\node [style=none] (16) at (1.5, 0.75) {$g_{Lin}$};
		\node [style=none] (17) at (-0.75, -1.5) {};
		\node [style=none] (18) at (0.75, -1.5) {};
		\node [style=none] (19) at (-0.75, -0.5) {};
		\node [style=none] (20) at (0, -1) {$g_{Lin}$};
		\node [style=none] (21) at (0.75, -0.5) {};
		\node [style=none] (22) at (1.25, 1.25) {};
		\node [style=none] (23) at (1.75, 1.25) {};
		\node [style=none] (24) at (0.25, -0.5) {};
		\node [style=none] (25) at (-0.25, -0.5) {};
		\node [style=none] (26) at (1.5, 0.25) {};
		\node [style=none] (27) at (-3, 2) {};
		\node [style=none] (28) at (0, 2) {};
		\node [style=none] (29) at (3, 2) {};
		\node [style=none] (30) at (0, -3.5) {};
		\node [style=none] (31) at (0, -1.5) {};
		\node [align=left, style=none] (32) at (3, -0.25) {$\vect{p_1} = g_{Lin}(\langvec{breathe}, \langvec{fire})$};
		\node [align=left, style=none] (33) at (2.5, -2.5) {$\vect{p_2} = g_{Lin}(\langvec{dragons}, \vect{p_1})$};
	\end{pgfonlayer}
	\begin{pgfonlayer}{edgelayer}
		\draw (0.center) to (1.center);
		\draw (1.center) to (2.center);
		\draw (0.center) to (2.center);
		\draw (3.center) to (4.center);
		\draw (3.center) to (5.center);
		\draw (4.center) to (5.center);
		\draw (6.center) to (7.center);
		\draw (6.center) to (8.center);
		\draw (7.center) to (8.center);
		\draw (12.center) to (14.center);
		\draw (14.center) to (15.center);
		\draw (15.center) to (13.center);
		\draw (13.center) to (12.center);
		\draw (19.center) to (17.center);
		\draw (17.center) to (18.center);
		\draw (18.center) to (21.center);
		\draw (21.center) to (19.center);
		\draw (29.center) to (23.center);
		\draw (28.center) to (22.center);
		\draw (26.center) to (24.center);
		\draw (27.center) to (25.center);
		\draw (31.center) to (30.center);
	\end{pgfonlayer}
\end{tikzpicture}

%% file: RNN2.tikz
\begin{tikzpicture}[{every path/.style}=thick]
	\begin{pgfonlayer}{nodelayer}
		\node [style=none] (0) at (-5.5, 0.75) {};
		\node [style=none] (1) at (-6.5, 0) {};
		\node [style=none] (2) at (-4.5, 0) {};
		\node [style=none] (3) at (0, 3.5) {};
		\node [style=none] (4) at (-0.5, 3) {};
		\node [style=none] (5) at (0.5, 3) {};
		\node [style=none] (6) at (5.5, 0.75) {};
		\node [style=none] (7) at (4.5, 0) {};
		\node [style=none] (8) at (6.5, 0) {};
		\node [style=none] (9) at (-5.5, 1.25) {\lang{dragons}};
		\node [style=none] (10) at (0, 4.5) {\lang{breathe}};
		\node [style=none] (11) at (5.5, 1.25) {\lang{fire}};
		\node [style=none] (12) at (-0.5, 2.25) {};
		\node [style=none] (13) at (1, 2.25) {};
		\node [style=none] (14) at (-0.5, 1.5) {};
		\node [style=none] (15) at (1, 1.5) {};
		\node [style=none] (16) at (0.25, 1.875) {$g_{Lin}$};
		\node [style=none] (17) at (-0.75, 0.25) {};
		\node [style=none] (18) at (0.75, 0.25) {};
		\node [style=none] (19) at (-0.75, 1) {};
		\node [style=none] (20) at (0, 0.625) {$g_{Lin}$};
		\node [style=none] (21) at (0.75, 1) {};
		\node [style=none] (22) at (0, 2.25) {};
		\node [style=none] (23) at (0.5, 2.25) {};
		\node [style=none] (24) at (0.25, 1) {};
		\node [style=none] (25) at (-0.25, 1) {};
		\node [style=none] (26) at (0.25, 1.5) {};
		\node [style=none] (27) at (-5.5, -0.25) {};
		\node [style=none] (28) at (0, 3) {};
		\node [style=none] (29) at (5.5, -0.25) {};
		\node [style=none] (30) at (0, -1.75) {};
		\node [style=none] (31) at (0, 0.25) {};
		\node [style=none] (32) at (1.5, 2.25) {};
		\node [style=none] (33) at (1.5, -0.25) {};
		\node [style=none] (34) at (-1.5, 1) {};
		\node [style=none] (35) at (-1.5, -0.25) {};
		\node [style=none] (36) at (0, 4) {};
		\node [style=none] (37) at (-4, 0) {};
		\node [style=none] (38) at (4, 0) {};
		\node [style=none] (39) at (-5.5, 0) {};
		\node [style=none] (40) at (5.5, 0) {};
	\end{pgfonlayer}
	\begin{pgfonlayer}{edgelayer}
		\draw (0.center) to (1.center);
		\draw (1.center) to (2.center);
		\draw (0.center) to (2.center);
		\draw (3.center) to (4.center);
		\draw (3.center) to (5.center);
		\draw (4.center) to (5.center);
		\draw (6.center) to (7.center);
		\draw (6.center) to (8.center);
		\draw (7.center) to (8.center);
		\draw (12.center) to (14.center);
		\draw (14.center) to (15.center);
		\draw (15.center) to (13.center);
		\draw (13.center) to (12.center);
		\draw (19.center) to (17.center);
		\draw (17.center) to (18.center);
		\draw (18.center) to (21.center);
		\draw (21.center) to (19.center);
		\draw (28.center) to (22.center);
		\draw (26.center) to (24.center);
		\draw (31.center) to (30.center);
		\draw [bend left=90, looseness=1.00] (23.center) to (32.center);
		\draw (32.center) to (33.center);
		\draw [bend right=90, looseness=1.25] (25.center) to (34.center);
		\draw (34.center) to (35.center);
		\draw (36.center) to (37.center);
		\draw (36.center) to (38.center);
		\draw (37.center) to (38.center);
		\draw [bend right=90, looseness=0.75] (27.center) to (35.center);
		\draw [bend right=90, looseness=0.75] (33.center) to (29.center);
		\draw (39.center) to (27.center);
		\draw (40.center) to (29.center);
	\end{pgfonlayer}
\end{tikzpicture}

%% file: RNNA.tikz
\begin{tikzpicture}[{every path/.style}=thick]
	\begin{pgfonlayer}{nodelayer}
		\node [style=none] (0) at (0, 2.5) {};
		\node [style=none] (1) at (0.5, 2) {};
		\node [style=none] (2) at (-0.5, 2) {};
		\node [style=none] (3) at (0, 0.25) {};
		\node [style=none] (4) at (-1.5, 0.25) {};
		\node [style=none] (5) at (0, 1) {};
		\node [style=none] (6) at (-0.75, 0.5) {$g_{Lin}$};
		\node [style=none] (7) at (-1.5, 1) {};
		\node [style=none] (8) at (-1, 1) {};
		\node [style=none] (9) at (-0.5, 1) {};
		\node [style=none] (10) at (0, 2) {};
		\node [style=none] (11) at (-0.75, -0.75) {};
		\node [style=none] (12) at (-0.75, 0.25) {};
		\node [style=none] (13) at (0.75, 1) {};
		\node [style=none] (14) at (0.75, -0.75) {};
		\node [style=none] (15) at (0, 3) {};
		\node [style=none] (16) at (2.75, 0) {};
		\node [style=none] (17) at (-2.75, 0) {};
	\end{pgfonlayer}
	\begin{pgfonlayer}{edgelayer}
		\draw (0.center) to (1.center);
		\draw (0.center) to (2.center);
		\draw (1.center) to (2.center);
		\draw (5.center) to (3.center);
		\draw (3.center) to (4.center);
		\draw (4.center) to (7.center);
		\draw (7.center) to (5.center);
		\draw (12.center) to (11.center);
		\draw [bend left=90, looseness=1.25] (9.center) to (13.center);
		\draw (13.center) to (14.center);
		\draw (15.center) to (16.center);
		\draw (15.center) to (17.center);
		\draw (16.center) to (17.center);
		\draw [in=90, out=-90, looseness=1.25] (10.center) to (8.center);
	\end{pgfonlayer}
\end{tikzpicture}

%% file: RNNIV.tikz
\begin{tikzpicture}[{every path/.style}=thick]
	\begin{pgfonlayer}{nodelayer}
		\node [style=none] (0) at (0, 2.5) {};
		\node [style=none] (1) at (0.5, 2) {};
		\node [style=none] (2) at (-0.5, 2) {};
		\node [style=none] (3) at (1.5, 1) {};
		\node [style=none] (4) at (0, 1) {};
		\node [style=none] (5) at (1.5, 0.25) {};
		\node [style=none] (6) at (0, 0.25) {};
		\node [style=none] (7) at (0.75, 0.75) {$g_{Lin}$};
		\node [style=none] (8) at (1, 1) {};
		\node [style=none] (9) at (0.5, 1) {};
		\node [style=none] (10) at (0.75, -1) {};
		\node [style=none] (11) at (0.75, 0.25) {};
		\node [style=none] (12) at (0, 2) {};
		\node [style=none] (13) at (-0.75, 1) {};
		\node [style=none] (14) at (-0.75, -1) {};
		\node [style=none] (15) at (0, 3) {};
		\node [style=none] (16) at (2.75, 0) {};
		\node [style=none] (17) at (-2.75, 0) {};
	\end{pgfonlayer}
	\begin{pgfonlayer}{edgelayer}
		\draw (0.center) to (1.center);
		\draw (0.center) to (2.center);
		\draw (1.center) to (2.center);
		\draw (3.center) to (5.center);
		\draw (5.center) to (6.center);
		\draw (6.center) to (4.center);
		\draw (4.center) to (3.center);
		\draw [in=90, out=-90, looseness=1.25] (12.center) to (8.center);
		\draw (11.center) to (10.center);
		\draw [bend right=90, looseness=1.25] (9.center) to (13.center);
		\draw (13.center) to (14.center);
		\draw (15.center) to (16.center);
		\draw (15.center) to (17.center);
		\draw (16.center) to (17.center);
	\end{pgfonlayer}
\end{tikzpicture}

%% file: relpron.tikz
\begin{tikzpicture}[{every path/.style}=thick]
	\begin{pgfonlayer}{nodelayer}
		\node [style=none, anchor=mid] (0) at (-1.75, 1.25) {\lang{dragons}};
		\node [style=none, anchor=mid] (1) at (2.5, 1.25) {\lang{breathe}};
		\node [style=none, anchor=mid] (2) at (4.25, 1.25) {\lang{fire}};
		\node [style=none] (3) at (-1.75, 0) {};
		\node [style=none] (4) at (-1, 0) {};
		\node [style=none] (5) at (2.5, 0) {};
		\node [style=none] (6) at (3, 0) {};
		\node [style=none] (7) at (4.25, 0) {};
		\node [style=none] (8) at (2.5, -0.5) {};
		\node [style=none] (9) at (4.75, 0) {};
		\node [style=none] (10) at (3.75, 0) {};
		\node [style=none] (11) at (4.25, 0.5) {};
		\node [style=none] (12) at (-1.75, 0.5) {};
		\node [style=none] (13) at (-2.25, 0) {};
		\node [style=none] (14) at (-1.25, 0) {};
		\node [style=none] (15) at (2.5, 0.75) {};
		\node [style=none] (16) at (1.5, 0) {};
		\node [style=none] (17) at (3.5, 0) {};
		\node [style={small_circle}] (18) at (0, 0) {};
		\node [style={small_circle}] (19) at (0.75, 0.25) {};
		\node [style=none] (20) at (1.25, 0) {};
		\node [style=none] (21) at (0, -1.5) {};
		\node [style=none] (22) at (-1, 0.5) {};
		\node [style=none] (23) at (-0.5, 0.5) {};
		\node [style=none] (24) at (0.5, 0.5) {};
		\node [style=none] (25) at (1.25, 0.5) {};
		\node [style=none] (26) at (2, 0) {};
		\node [style=none] (27) at (0.75, -0.5) {};
		\node [style=none] (28) at (0, 1.25) {\lang{who}};
	\end{pgfonlayer}
	\begin{pgfonlayer}{edgelayer}
		\draw [bend right=90, looseness=1.25] (3.center) to (4.center);
		\draw [bend right=90, looseness=1.00] (6.center) to (7.center);
		\draw (5.center) to (8.center);
		\draw [style=swap] (11.center) to (9.center);
		\draw [style=swap] (9.center) to (10.center);
		\draw [style=swap] (10.center) to (11.center);
		\draw [style=swap] (12.center) to (14.center);
		\draw [style=swap] (14.center) to (13.center);
		\draw [style=swap] (13.center) to (12.center);
		\draw [style=swap] (15.center) to (17.center);
		\draw [style=swap] (17.center) to (16.center);
		\draw [style=swap] (16.center) to (15.center);
		\draw (18) to (21.center);
		\draw (22.center) to (4.center);
		\draw [bend left=90, looseness=1.25] (22.center) to (23.center);
		\draw [bend left=90, looseness=1.00] (24.center) to (25.center);
		\draw [in=180, out=-90, looseness=1.25] (23.center) to (18);
		\draw [in=0, out=-90, looseness=1.25] (24.center) to (18);
		\draw (25.center) to (20.center);
		\draw [bend left=90, looseness=1.25] (26.center) to (20.center);
		\draw (19) to (27.center);
		\draw [bend right=90, looseness=0.75] (27.center) to (8.center);
	\end{pgfonlayer}
\end{tikzpicture}

%% file: relpronyanked.tikz
\begin{tikzpicture}[{every path/.style}=thick]
	\begin{pgfonlayer}{nodelayer}
		\node [style=none, anchor=mid] (0) at (-0.75, 1.25) {\lang{dragons}};
		\node [style=none, anchor=mid] (1) at (1.25, 1.25) {\lang{breathe}};
		\node [style=none, anchor=mid] (2) at (3, 1.25) {\lang{fire}};
		\node [style=none] (3) at (-0.75, 0) {};
		\node [style=none] (4) at (1.25, 0) {};
		\node [style=none] (5) at (1.75, 0) {};
		\node [style=none] (6) at (3, 0) {};
		\node [style={small_circle}] (7) at (1.25, -1) {};
		\node [style=none] (8) at (3.5, 0) {};
		\node [style=none] (9) at (2.5, 0) {};
		\node [style=none] (10) at (3, 0.5) {};
		\node [style=none] (11) at (-0.75, 0.5) {};
		\node [style=none] (12) at (-1.25, 0) {};
		\node [style=none] (13) at (-0.25, 0) {};
		\node [style=none] (14) at (1.25, 0.75) {};
		\node [style=none] (15) at (0.25, 0) {};
		\node [style=none] (16) at (2.25, 0) {};
		\node [style={small_circle}] (17) at (0, -0.5) {};
		\node [style=none] (18) at (0, -1.5) {};
		\node [style=none] (19) at (0.75, 0) {};
	\end{pgfonlayer}
	\begin{pgfonlayer}{edgelayer}
		\draw [bend right=90, looseness=1.25] (5.center) to (6.center);
		\draw (4.center) to (7);
		\draw [style=swap] (10.center) to (8.center);
		\draw [style=swap] (8.center) to (9.center);
		\draw [style=swap] (9.center) to (10.center);
		\draw [style=swap] (11.center) to (13.center);
		\draw [style=swap] (13.center) to (12.center);
		\draw [style=swap] (12.center) to (11.center);
		\draw [style=swap] (14.center) to (16.center);
		\draw [style=swap] (16.center) to (15.center);
		\draw [style=swap] (15.center) to (14.center);
		\draw (17) to (18.center);
		\draw [in=180, out=-90, looseness=1.25] (3.center) to (17);
		\draw [in=0, out=-90, looseness=1.25] (19.center) to (17);
	\end{pgfonlayer}
\end{tikzpicture}

%% file: reflexivepron.tikz
\begin{tikzpicture}[{every path/.style}=thick]
	\begin{pgfonlayer}{nodelayer}
		\node [style=none, anchor=mid] (0) at (-1.75, 1.5) {\lang{John}};
		\node [style=none, anchor=mid] (1) at (0, 1.5) {\lang{loves}};
		\node [style=none, anchor=mid] (2) at (2.5, 1.5) {\lang{himself}};
		\node [style=none] (3) at (-1.75, 0) {};
		\node [style=none] (4) at (0, 0) {};
		\node [style=none] (5) at (0.5, 0) {};
		\node [style=none] (6) at (2.5, 0) {};
		\node [style=none] (7) at (1.5, 0) {};
		\node [style=none] (8) at (-1.75, 0.5) {};
		\node [style=none] (9) at (-2.25, 0) {};
		\node [style=none] (10) at (-1.25, 0) {};
		\node [style=none] (11) at (0, 0.75) {};
		\node [style=none] (12) at (-1, 0) {};
		\node [style=none] (13) at (1, 0) {};
		\node [style=none] (14) at (-0.5, 0) {};
		\node [style=none] (15) at (2, 0) {};
		\node [style=none] (16) at (3, 0) {};
		\node [style=none] (17) at (3.5, 0.5) {};
		\node [style=none] (18) at (3.5, -1.5) {};
		\node [style={small_circle}] (19) at (2.25, 0.5) {};
		\node [style=none] (20) at (3, 0.5) {};
		\node [style=none] (21) at (1.5, 0.25) {};
		\node [style=none] (22) at (2, 0.25) {};
	\end{pgfonlayer}
	\begin{pgfonlayer}{edgelayer}
		\draw [style=swap] (8.center) to (10.center);
		\draw [style=swap] (10.center) to (9.center);
		\draw [style=swap] (9.center) to (8.center);
		\draw [style=swap] (11.center) to (13.center);
		\draw [style=swap] (13.center) to (12.center);
		\draw [style=swap] (12.center) to (11.center);
		\draw [bend right=90, looseness=1.25] (5.center) to (7.center);
		\draw [bend right=90, looseness=1.25] (4.center) to (15.center);
		\draw [bend right=90, looseness=1.25] (14.center) to (6.center);
		\draw [bend right=90, looseness=1.00] (3.center) to (16.center);
		\draw (17.center) to (18.center);
		\draw (16.center) to (20.center);
		\draw [bend right=90, looseness=1.50] (20.center) to (19);
		\draw [in=90, out=0, looseness=1.00] (19) to (6.center);
		\draw [in=-90, out=90, looseness=0.50] (15.center) to (21.center);
		\draw [bend left=90, looseness=1.50] (21.center) to (17.center);
		\draw [in=90, out=180, looseness=1.00] (19) to (22.center);
		\draw [in=90, out=-90, looseness=0.50] (22.center) to (7.center);
	\end{pgfonlayer}
\end{tikzpicture}

%% file: reflexivepronyanked.tikz
\begin{tikzpicture}
	\begin{pgfonlayer}{nodelayer}
		\node [style=none] (0) at (-2.25, 0) {};
		\node [style={small_circle}] (1) at (-1.75, -0.5) {};
		\node [style=none] (2) at (-0.5, -1) {};
		\node [style=none] (3) at (-0.5, 0) {};
		\node [style=none] (4) at (-1.75, 0) {};
		\node [style=none] (5) at (0, -0.5) {};
		\node [style=none] (6) at (-1.75, 0.5) {};
		\node [style=none, anchor=mid] (7) at (-1.75, 1.5) {\lang{John}};
		\node [style=none] (8) at (0, 0) {};
		\node [style=none, anchor=mid] (9) at (0, 1.5) {\lang{loves}};
		\node [style=none] (10) at (0.5, 0) {};
		\node [style=none] (11) at (-1.25, 0) {};
		\node [style=none] (12) at (-1, 0) {};
		\node [style=none] (13) at (-2.25, -1) {};
		\node [style=none] (14) at (1, 0) {};
		\node [style=none] (15) at (0, 0.75) {};
		\node [style=none] (16) at (0.5, -0.5) {};
		\node [style=none] (17) at (-1.25, -1) {};
		\node [style=none] (18) at (0, -1) {};
		\node [style=none] (19) at (0.5, -1) {};
		\node [style=none] (20) at (0.5, -1.5) {};
	\end{pgfonlayer}
	\begin{pgfonlayer}{edgelayer}
		\draw [style=swap] (6.center) to (11.center);
		\draw [style=swap] (11.center) to (0.center);
		\draw [style=swap] (0.center) to (6.center);
		\draw [style=swap] (15.center) to (14.center);
		\draw [style=swap] (14.center) to (12.center);
		\draw [style=swap] (12.center) to (15.center);
		\draw [in=90, out=180, looseness=1.00] (1) to (13.center);
		\draw [in=90, out=0, looseness=1.25] (1) to (17.center);
		\draw (4.center) to (1);
		\draw (3.center) to (2.center);
		\draw [bend right=90, looseness=1.25] (17.center) to (2.center);
		\draw (8.center) to (5.center);
		\draw (10.center) to (16.center);
		\draw [in=90, out=-90, looseness=0.75] (16.center) to (18.center);
		\draw [in=90, out=-90, looseness=0.75] (5.center) to (19.center);
		\draw [bend right=90, looseness=1.00] (13.center) to (18.center);
		\draw (19.center) to (20.center);
	\end{pgfonlayer}
\end{tikzpicture}

%% file: CatsvRNNsNeSyIfCoLog.bbl
\begin{thebibliography}{17}
\providecommand{\natexlab}[1]{#1}
\providecommand{\url}[1]{\texttt{#1}}
\expandafter\ifx\csname urlstyle\endcsname\relax
  \providecommand{\doi}[1]{doi: #1}\else
  \providecommand{\doi}{doi: \begingroup \urlstyle{rm}\Url}\fi

\bibitem[Baroni and Zamparelli(2010)]{BaroniZamparelli2010}
Marco Baroni and Roberto Zamparelli.
\newblock Nouns are vectors, adjectives are matrices: Representing
  adjective-noun constructions in semantic space.
\newblock In \emph{Proceedings of the 2010 Conference on Empirical Methods in
  Natural Language Processing}, pages 1183--1193. Association for Computational
  Linguistics, 2010.

\bibitem[Bolt et~al.(2017)Bolt, Coecke, Genovese, Lewis, Marsden, and
  Piedeleu]{bolt2017}
Joe Bolt, Bob Coecke, Fabrizio Genovese, Martha Lewis, Dan Marsden, and Robin
  Piedeleu.
\newblock Interacting conceptual spaces i: Grammatical composition of concepts.
\newblock \emph{arXiv preprint arXiv:1703.08314}, 2017.

\bibitem[Bowman and Potts(2015)]{BowmanPotts2015}
Samuel~R Bowman and Christopher Potts.
\newblock Recursive neural networks can learn logical semantics.
\newblock \emph{ACL-IJCNLP 2015}, page~12, 2015.

\bibitem[Bullinaria and Levy(2007)]{Bullinaria2007}
J.A. Bullinaria and J.P. Levy.
\newblock Extracting semantic representations from word co-occurrence
  statistics: A computational study.
\newblock \emph{Behavior research methods}, 39\penalty0 (3):\penalty0 510--526,
  2007.

\bibitem[Coecke(2013)]{Coecke2013}
B.~Coecke.
\newblock An alternative gospel of structure: order, composition, processes.
\newblock In C.~Heunen, M.~Sadrzadeh, and E.~Grefenstette, editors,
  \emph{Quantum Physics and Linguistics. A Compositional, Diagrammatic
  Discourse}, pages 1--22. Oxford University Press, 2013.

\bibitem[Coecke and Paquette(2011)]{CoeckePaquette2011}
B.~Coecke and E.O. Paquette.
\newblock Categories for the practising physicist.
\newblock In \emph{New Structures for Physics}, pages 173--286. Springer, 2011.
\newblock \doi{10.1007/978-3-642-12821-9}.

\bibitem[Coecke et~al.(2010)Coecke, Sadrzadeh, and
  Clark]{CoeckeSadrzadehClark2010}
B.~Coecke, M.~Sadrzadeh, and S.~Clark.
\newblock Mathematical foundations for a compositional distributional model of
  meaning.
\newblock \emph{Linguistic Analysis}, 36:\penalty0 345--384, 2010.

\bibitem[Grefenstette and Sadrzadeh(2011)]{GrefenstetteSadrzadeh2011}
Edward Grefenstette and Mehrnoosh Sadrzadeh.
\newblock Experimental support for a categorical compositional distributional
  model of meaning.
\newblock In \emph{Proceedings of the Conference on Empirical Methods in
  Natural Language Processing}, pages 1394--1404. Association for Computational
  Linguistics, 2011.

\bibitem[Grefenstette et~al.(2013)Grefenstette, Dinu, Zhang, Sadrzadeh, and
  Baroni]{GrefenstetteDinuZhangSadrzadehBaroni2013}
Edward Grefenstette, Georgiana Dinu, Yao-Zhong Zhang, Mehrnoosh Sadrzadeh, and
  Marco Baroni.
\newblock Multi-step regression learning for compositional distributional
  semantics.
\newblock \emph{arXiv preprint arXiv:1301.6939}, 2013.

\bibitem[Kartsaklis et~al.(2012)Kartsaklis, Sadrzadeh, and
  Pulman]{KartsaklisSadrzadehPulman2012}
Dimitri Kartsaklis, Mehrnoosh Sadrzadeh, and Stephen Pulman.
\newblock A unified sentence space for categorical distributional-compositional
  semantics: Theory and experiments.
\newblock In \emph{Proceedings of COLING 2012: Posters}, pages 549--558, 2012.

\bibitem[Maillard et~al.(2014)Maillard, Clark, and Grefenstette]{maillard2014}
Jean Maillard, Stephen Clark, and Edward Grefenstette.
\newblock A type-driven tensor-based semantics for ccg.
\newblock \emph{EACL 2014}, page~46, 2014.

\bibitem[Mikolov et~al.(2013)Mikolov, Sutskever, Chen, Corrado, and
  Dean]{mikolov2013}
Tomas Mikolov, Ilya Sutskever, Kai Chen, Greg~S Corrado, and Jeff Dean.
\newblock Distributed representations of words and phrases and their
  compositionality.
\newblock In \emph{Advances in neural information processing systems}, pages
  3111--3119, 2013.

\bibitem[Mitchell and Lapata(2010)]{Mitchell2010}
J.~Mitchell and M.~Lapata.
\newblock Composition in distributional models of semantics.
\newblock \emph{Cognitive science}, 34\penalty0 (8):\penalty0 1388--1429, 2010.

\bibitem[Paperno et~al.(2014)Paperno, Baroni, et~al.]{paperno2014}
Denis Paperno, Marco Baroni, et~al.
\newblock A practical and linguistically-motivated approach to compositional
  distributional semantics.
\newblock In \emph{Proceedings of the 52nd Annual Meeting of the Association
  for Computational Linguistics (Volume 1: Long Papers)}, volume~1, pages
  90--99, 2014.

\bibitem[Preller and Sadrzadeh(2011)]{PrellerSadrzadeh2011}
A.~Preller and M.~Sadrzadeh.
\newblock Bell states and negative sentences in the distributed model of
  meaning.
\newblock \emph{Electronic Notes in Theoretical Computer Science}, 270\penalty0
  (2):\penalty0 141--153, 2011.
\newblock \doi{10.1016/j.entcs.2011.01.028}.

\bibitem[Sadrzadeh et~al.(2013)Sadrzadeh, Clark, and Coecke]{FrobMeanI}
M.~Sadrzadeh, S.~Clark, and B.~Coecke.
\newblock The {F}robenius anatomy of word meanings {I}: subject and object
  relative pronouns.
\newblock \emph{Journal of Logic and Computation}, 23:\penalty0 1293--1317,
  2013.
\newblock ar{X}iv:1404.5278.

\bibitem[Socher et~al.(2013)Socher, Perelygin, Wu, Chuang, Manning, Ng, and
  Potts]{SocherPerelyginWuChuangManningNgPotts2013}
Richard Socher, Alex Perelygin, Jean Wu, Jason Chuang, Christopher~D Manning,
  Andrew Ng, and Christopher Potts.
\newblock Recursive deep models for semantic compositionality over a sentiment
  treebank.
\newblock In \emph{Proceedings of the 2013 conference on empirical methods in
  natural language processing}, pages 1631--1642, 2013.

\end{thebibliography}
